# Mitochondria-based Renal Cell Carcinoma Subtyping: Learning from Deep vs. Flat Feature Representations

**Peter J. Schüffler**                                    schueffp@mskcc.org
*Department of Medical Physics*
*Memorial Sloan Kettering Cancer Center*
*New York City, NY, USA*

**Judy Sarungbam**                                    sarungbj@mskcc.org
*Department of Pathology*
*Memorial Sloan Kettering Cancer Center*
*New York City, NY, USA*

**Hassan Muhammad**                          ham2024@med.cornell.edu
*Weill Cornell Medicine*
*New York City, NY, USA*

**Ed Reznik**                                    reznike@mskcc.org
*Computational Biology Program & Center for Molecular Oncology*
*Memorial Sloan Kettering Cancer Center*
*New York City, NY, USA*

**Satish K. Tickoo**                                    tickoos@mskcc.org
*Department of Pathology*
*Memorial Sloan Kettering Cancer Center*
*New York City, NY, USA*

**Thomas J. Fuchs**                                    fuchst@mskcc.org
*Department of Medical Physics*
*Memorial Sloan Kettering Cancer Center*
*New York City, NY, USA*

## Abstract

Accurate subtyping of renal cell carcinoma (RCC) is of crucial importance for understanding disease progression and for making informed treatment decisions. New discoveries of significant alterations to mitochondria between subtypes make immunohistochemical (IHC) staining based image classification an imperative.

Until now, accurate quantification and subtyping was made impossible by huge IHC variations, the absence of cell membrane staining for cytoplasm segmentation as well as the complete lack of systems for robust and reproducible image based classification.

In this paper we present a comprehensive classification framework to overcome these challenges for tissue microarrays (TMA) of RCCs. We compare and evaluate models based on domain specific hand-crafted "flat"-features versus "deep" feature representations from various layers of a pre-trained convolutional neural network (CNN). The best model reaches a cross-validation accuracy of 89%, which demonstrates for the first time, that robust mitochondria-based subtyping of renal cancer is feasible.

## Introduction

Renal cell carcinoma (RCC) arises in several subtypes that differ in histology, morphology and genetics. Clear cell (CC) RCC is the most common subtype, while relatively rare types include clear cell papillary (CCP) RCC, TCEB1-mutated RCC, and to a lesser extent MiTF /translocation family RCC and Tuberous sclerosis-associated RCC (Hakimi et al. (2015); Rohan et al. (2011); Argani et al. (2001)). These subtypes need to be differentiated from each other because of their different prognostic implications and different mechanisms for tumorigenesis.

Recent work by us (Reznik et al. (2016)) and others (Meierhofer et al. (2004); Nilsson et al. (2015); Davis et al. (2014)) has consistently shown that several subtypes of RCC show drastic changes in the number and activity of mitochondria. Alteration in mitochondrial copy number may play a significant role in cancer biology. CC RCC is for example genetically driven by the activation of hypoxia-inducible factor *HIF*, a transcription factor that activates the cellular response to hypoxic stress. As a result, CC RCC is well-known to down-regulate the expression of metabolic and mitochondrial genes relative to normal tissue (Tickoo et al. (2000); The Cancer Genome Atlas Research Network (2013)).

In stark contrast, renal oncocytoma (ONC), a more rare benign tumor of the kidney, is uniquely characterized by the massive accumulation of mitochondria in the cytoplasm (Joshi et al. (2015)). Recent work has demonstrated that these mitochondria harbor mutations in their DNA (mtDNA), and that mitochondrial accumulation in ONC arises from defects in autophagy, which would normally clear defective mitochondria from the cell.

Notably, many rare subtypes of RCC remain genetically uncharacterized. Therefore, we examine in this study the histologic differentiation of ONC, CC, and CCP, which bears morphologic features resembling both the clear-cell and papillary subtypes of RCC (Rohan et al. (2011)). While CCP RCCs are distinguished from other RCCs by immunohisto-chemical staining patterns, disambiguation of CC from CCP tumors remains challenging (Sarungbam et al. (2016)).
For a systematic, objective and automated differentiation of these tumor subtypes, we use techniques drawn from computational pathology, which have shown promising predictive power in earlier studies.

## Previous Work

RCC subtyping has been investigated previously on genetic profiles (e.g. Brannon et al. (2010)), tissue microarray readouts (e.g. Shi et al. (2004)), or H&E stained images features (e.g. Chaudry et al. (2009); Raza et al. (2009); Yeh et al. (2014)). To the best of our knowledge, there exists no subtyping based on mitochondria histology images.

Fuchs and Buhmann (2011) introduced the field of computational pathology offering advanced pipelines for automated classification of histology images. Deep learning with convolutional neural networks (CNNs) gain more and more impact in this scenario. Their applications range from cell detection and classification (Cireşan et al. (2013); Wang et al.

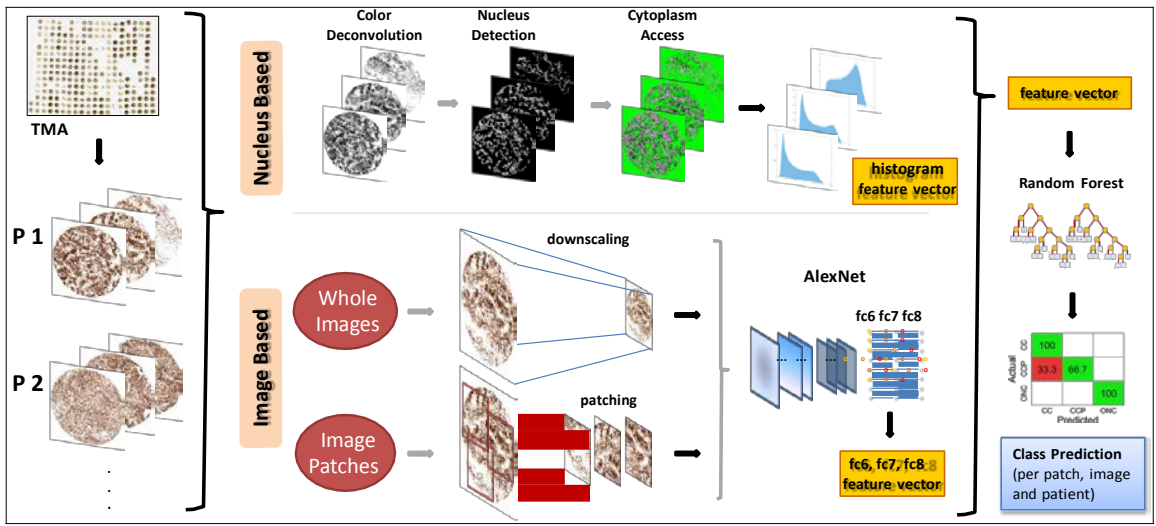

Figure 1: **Scheme of the mitochondria based RCC subtype classification. Top row:** in a classical structured approach, the cells are detected in the images to access their cytoplasm intensity histograms as feature vectors. **Bottom row:** In contrast to that complicated multistep approach, we test state-of-the-art CNN models on whole images and image patches for feature extraction from AlexNet's fully connected layers *fc6*, *fc7* and *fc8*. Feature vectors are then fed to the same cross-validation procedure for predicting CC, CCP or ONC subtypes.

(2014); Veta et al. (2015); Cruz-Roa et al. (2013); Giusti et al. (2014)) to overall tissue segmentation (Le et al. (2012)). CNNs usually need large data sets to learn from, which is why recent approaches make use of existing models pre-trained on large image databases (Bar et al. (2015)). These trained models can then further be fine-tuned to a given dataset for better fitting. CNNs usually show high performance compared to other methods in histology image computation (e.g. CNNs won the MICCAI 2013 grand challenge (Cireşan et al. (2013)) and the AMIDA13 contest (Veta et al. (2015)) on mitosis detection). Clear advantages of CNNs are their simple usability and the ability to unsupervised self-organization (Fukushima (1980)). However, the higher level features they learn are usually difficult to interpret (Yosinski et al. (2015)). As a recent extension of CNNs, the hybridization with random forests gains more and more attention in the community (Kontschieder et al. (2015); Long et al. (2015)) and has already shown promising results in medical image processing (Shah et al. (2016)). We contribute to this possibility by the use of native CNN features in random forest classifiers for advanced RCC subtyping.

## Contributions

We present in this study a new approach to extract high performing CNN features from histology images of a RCC tissue microarray (TMA) for subsequent subtype classification. We address several problems such as scaling and patching and compare the different network layers as feature vectors (c.f. Fig 1, bottom row). Further, we compare our method to a dedicated and more complex cytoplasm-based approach, which yields quantitative histogram features from the mitochondria, in contrast to the less intuitive CNN features (c.f. Fig 1, top row). We show that CNNs lack in information when they look at the downscaled whole images (and histogram information helps), but the CNN outperforms both methods when they look at dedicated image patches.

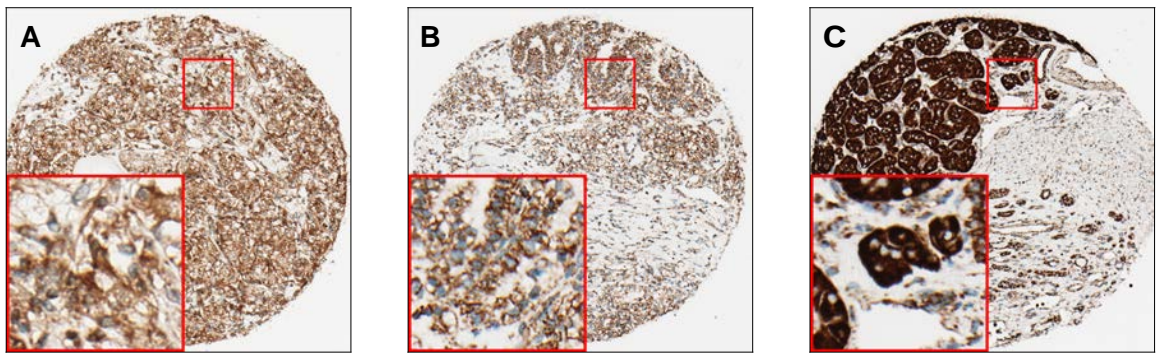

Figure 2: **Example image data.** Brownish *TOM20* staining indicates mitochondria in cytoplasm. **A:** CC type, **B:** CCP type, **C:** ONC type. ONC usually has extremely over-replicated mitochondria, whereas CC and CPP have an under-represented number. The automatic distinction of the three classes by mitochondria is a key task of this paper.

## Methods

75 TMA specimen replicates from 28 RCC patients (10 CC, 9 CCP, 9 ONC) were stained at the specific mitochondria receptor unit *TOM20*, resulting in brown color response. Cell nuclei were counter-stained with bluish *Eosine*. See Fig. 2 for three example images and S1[1] for a TMA overview. The TMA was scanned with an Aperio AT2 digital slide scanner at 20x magnification, resulting in single spot images of 1500x1500px size.

**Automatic White Balancing:** Each image has been corrected for scanning bias to get a neutral, white background. For this, each pixel was divided by the image's representative background color found by a window shifting algorithm over the image: each window $W$ is gray-scaled to $W_G$, and the normalized intensity histogram $I$ of $W_G$ is calculated. The entropy $H(I)$ is defined as:

$$H(I) = -\sum_i I_i \log(I_i)$$

where $I_i$ is the $i_{th}$ bin of $I$. The average color of the window $W$ with the smallest entropy $H(I)$ represents the background color.

---

1. Supplement available at end of document

## Cytoplasm Histogram Based Approach (*HIST*)

To quantify the mitochondria in the images, we describe their specific cytoplasm intensity histograms. The color deconvolution algorithm by Ruifrok and Johnston (2001) was used to separate bluish nuclei from mitochondria staining (Fig 3A-C). Using the complete mitochondria channel for histograms could exclude bright clear cells and potentially include non-cell structures in the histograms. Since we aim to include the information of all cancer cells in the image and not only stained ones, we suggest prior cell detection and cytoplasm segmentation.

Since the accurate cell segmentation of RCC can be very hard due to missing membrane information (and is not part of this study), we approximate the cytoplasm as rings around the nuclei. For this, the nucleus positions are determined on the nucleus channel by a watershed based algorithm introduced ealier (Schüffler et al. (2013)).

The cytoplasm on the mitochondria channel is then outlined as circular rings around the nuclei with thickness of 10px, as optically confirmed to fit our cells (Fig 3D, purple). RCC TMA images may be disturbed by tissue lesions in which the background is visible. Therefore, we use a background intensity threshold of $t = 220$ to exclude brighter image background pixels from the image and rings (c.f. Fig 3D, green). Therefore, the rings do not close properly, when the cells lie at a tissue border.

After this processing, the area on the rings defines the region of interest (ROI) of an image. The intensity histogram of the mitochondria channel in that ROI (c.f. Fig 3E) is the basis of the feature vector for that image. The feature vector *HIST* has a length of 517 and consists of a pyramid of normalized ROI histograms with 256, 128, 64, 32, 16, 8, and 4 bins, the four 256-bin histogram quantiles $q_1$, $q_2$, $q_3$ and $q_4$, the mean and median intensity, skewness, kurtosis, and a histogram-based H-score defined as:

$$HScore = 100 * \sum_{i=1}^{4} i * q_i$$

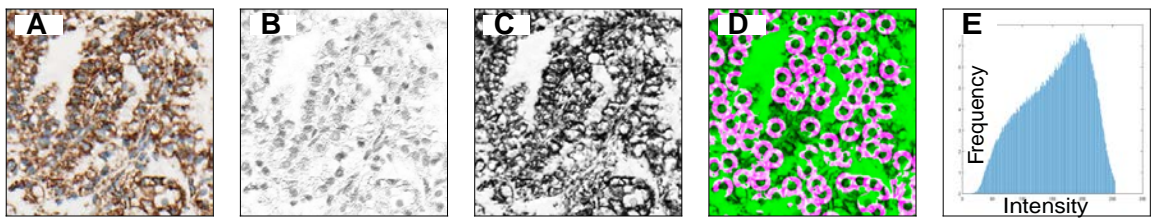

Figure 3: Processing of histology images. From the white-balanced image (**A** shows detail), color deconvolution reveals the nuclei (**B**) and mitochondria (**C**). The cytoplasm approximated as all rings around the nuclei (**D**, purple) defines the ROI for the histogram (**E**). A simple threshold excludes white background (D, green).

**Convolutional Neural Network**

CNNs have seen striking advancements over recent years in the area of medical image classification. To compare this state-of-the-art technique with the more complex nucleus-based approach, we employ the use of CNNs to deconstruct our TMA images into feature vectors using two different methodologies: (i) we apply CNNs on the whole TMA spots and (ii) we incorporate a patching technique to sample small areas from the whole image using a mask to ignore areas without relevant information.

AlexNet by Krizhevsky et al. (2012) is a well studied and popular CNN pre-trained on natural images. It offers a much simpler structure than other nets (e.g. GoogleNet) and can be used for our purposes easily. Due to the relatively low number of samples, training a CNN from scratch would not be feasible for our study. Also, as our main goal is to explore the power of naïve deep features, we do not fine-tune AlexNet for our purposes. Instead, we transfer the parameterization (tuning) step to random forests as primary classifiers.

The AlexNet architecture contains three fully connected layers ($fc6$ and $fc7$ with 4096 nodes (features), and $fc8$ before a final softmax classifier, with 1000 nodes (features)). Note that softmax normalizes $fc8$ to the probabilistic label vector of the CNN. Before that normalization, $fc8$ can be interpreted as an important feature vector for classification. Because each layer represents a different set of abstract features, it is important to choose one that best contributes to a well-performing classifier. Feature vectors were extracted from all three fully connected layers with the Caffe Deep Learning framework (Jia et al. (2014)) for CNN on a cluster of 20 NVIDIA TitanX GPUs.

Whole Image Approach (*CNN*)

A main advantage of CNNs is the simplicity of their application on raw images which is why we test our approach with whole images. Since RCC subtyping is rotation invariant, and to increase the dataset for CNNs, we extend the original data set for whole image feature extraction by rotating the original images by $90°$, $180°$ and $270°$ and flipping. The extended data set has $75 \cdot 4 \cdot 2 = 600$ images.

AlexNet has an input image size of 227x227px, and Caffe will downscale the whole image. This downscaling can cause loss of data due to a decrease in resolution. To explore this effect on classification performance, we introduce an image patching approach.

Image Patching Approach (*CNN$_p$*)

Each TMA spot was segmented into as many 227x227px partly overlapping patches that would fit within foreground areas of the TMA by a random sampling method according to following protocol (see Fig. 4):

To separate the background of an white-balanced image $I$, we defined a fixed threshold $t = 230$ which was applied on a Gaussian blurred ($\sigma = 2$), gray-scaled version of $I$, yielding the black/white foreground mask $BW$. To generate the patches, we first sampled 1000 random candidate patches of size 227x227px (AlexNet input size) uniformly over the whole image. A patch is then only accepted if:

- the corresponding foreground fills at least 80% of the patch (thus ignoring patches with large background content), and

- the patch has an overlap of the already patched area of at most 50% (thus reducing redundant patches, see Fig. 4), and

- the gray-scaled patch has an entropy $e \geq 4.6$, as empirically evaluated, excluding background structures not excluded by the background threshold (see Fig. 5).

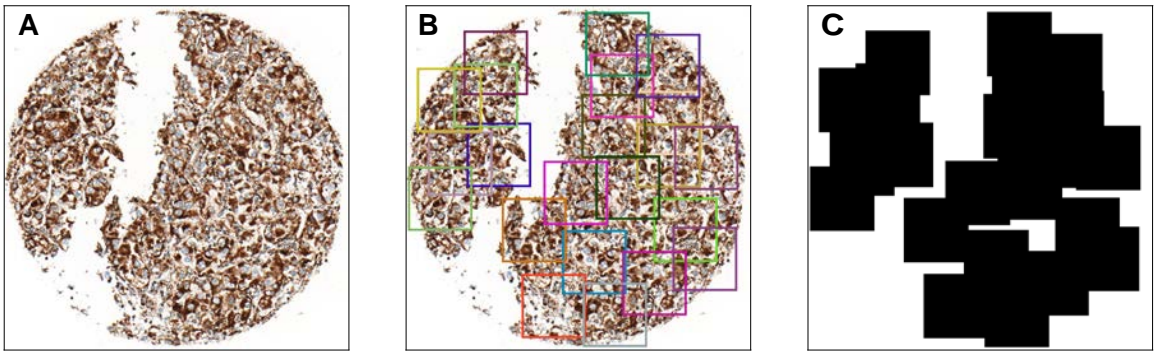

Figure 4: Example of patching. **A:** Original image. Candidate patches are randomly sampled at 227x227px (AlexNet input size). **B:** Accepted patches covered by less than 80% background and less than 50% already sampled area. **C:** Sampled area. Most of the foreground is covered.

Due to the high number of initial patch candidates, we cover as much as possible of the foreground area. Note that systematic sampling in a grid would include too many white patches, especially when the tissue is broken. On average, 29 patches were extracted per image (min 4, max 38 patches), or 76 patches per patient (min 7, max 106). In total, 1936 patches were sampled (see Fig. 6 for example patches). Patches were labeled with their parents' class, and feature vectors were extracted for each patch from $fc6$, $fc7$, $fc8$ analog to the whole image approach.

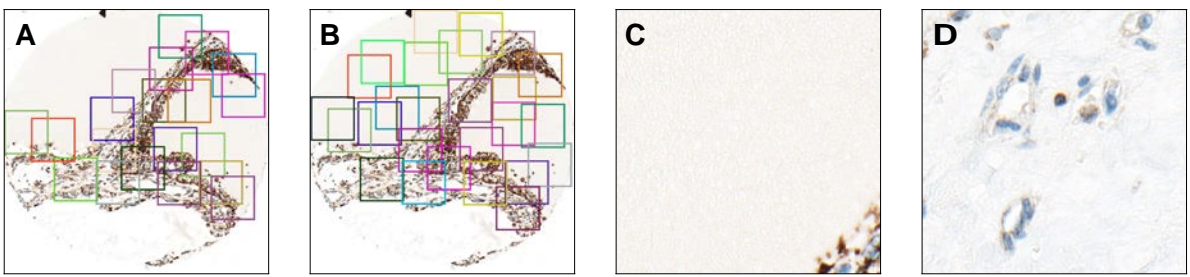

Figure 5: The entropy filter reduces the number of patches without content. **A:** Patching with entropy filter. **B:** Patching without entropy filter. **C:** Detail of a filtered patch with small entropy. **D:** Similar "background" as C, but higher entropy.

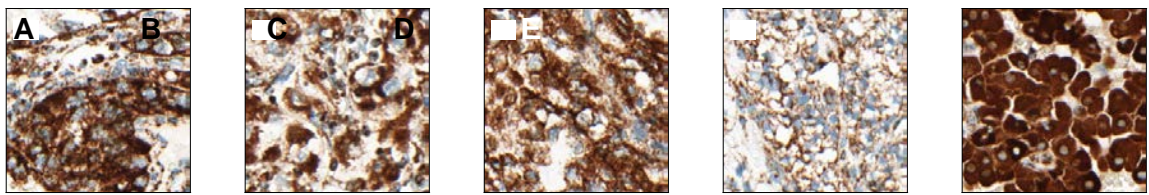

Figure 6: Examples of 227x227px image patches. **A+B:** CC cases, **C+D:** CCP cases, **E:** ONC case.

## Classification and LOPO Cross-Validation

To evaluate the classification performance of the proposed methods *HIST*, *CNN* and *CNN$_P$*, and their different feature vectors, we trained random forest models (Breiman (2001); Amit and Geman (1997)) with 50 trees on the generated feature vectors from each approach separately and combined to predict the cancer subtype CC, CCP or ONC. We chose 50 trees, as a larger number of trees did not show better performance (see S2). All other parameters (depth, number of features, etc) were not optimized and taken from the standards of the random forest implementation in R. Feature combination was done by concatenation of the vectors being simple and justified by the use of random forests, since they are not highly affected by different dynamic ranges or correlation.

Each classifier has been cross-validated in a leave-one-patient-out (LOPO-CV) scenario, training on the images of 27 patients and predicting the left-out patient's images. A patient's cancer subtype is determined as the class majority of his or her images. The multiple images per patient allow for an entropy-based confidence score for the patient's class.

## Results & Discussion

### Cytoplasm Intensity Histograms

The 75 cytoplasm intensity histograms of the TMA showed differential shape among the three cancer groups (see Fig. 7): while the CC types have a camel like shape (two peaks), the CCP cases show a dromedary like shape (one peak) and usually little dark pixels. The ONC subtypes, in contrast, have a clear main emphasis on dark pixels. By the histograms alone, one could conclude the CC cases are a mixture of CCP and ONC.

### Classification Performance

The mean error rate per class (**balanced error, BE**) is reported for patient classification into CC, CCP or ONC in Fig. 8. *HIST* has a BE of 21.5% in LOPO-CV (Fig. 8A). *CNN* reaches a BE of 27.8% (Fig. 8B) due to a higher number of misclassified CCs, which can be improved by adding *HIST* to 14.4%, (Fig. 8C). However, the best model is *CNN$_P$* (*fc*6) with a BE of 11.1% (Fig. 8D), only misclassifying CCP.

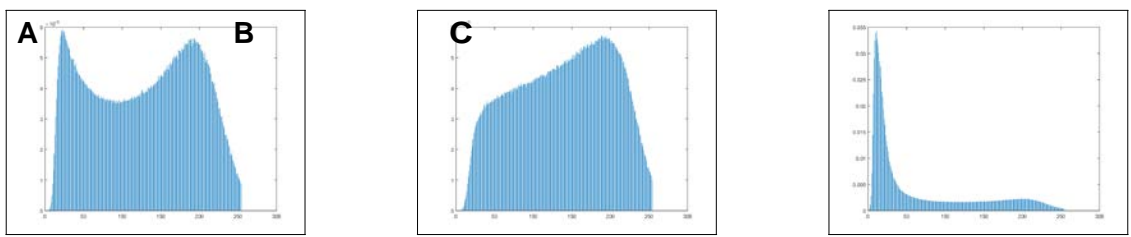

Figure 7: Mean cytoplasm intensity histograms of the three subtypes **A:** CC, **B:** CCP, **C:** ONC. X-Axis: intensity (0-255), y-axis: normalized counts.

An overview of all experiments with patient and image classification error is given in Fig. 9. It demonstrates that the histogram-based method (middle, blue) performs better than *CNN* on whole images, and can add to those features. But on the other hand, the superior models are *CNN_P*'s without histograms. It is remarkable that the *CNN* ($fc6+fc7+fc8$) and ($fc8$) show considerable performance of 25-35% classification error, although the images are highly scaled down. To test if the *CNN* still learns from tissue information, or if the mitochondrial average stain in the (downscaled) image is correlated with the class, we learned a baseline random forest on the average staining intensity per image. The performance of that classifier with per patient error 39% and per image error 43% is lower than that of the *CNN*, supporting the hypothesis that the network can learn from the scaled images.

Further, the patching method works significantly better with $fc6$ than e.g. for $fc6+fc7+fc8$. An explanation can be that the concatenated feature vector will be too large (9k) compared to the number of samples (2k patches or 600 images) considering the fact it is better on the many patches than on the few whole images.

In general, ONC is easier to separate than CC and CCP. This is supported by a clustering of the intensity histograms clearly separating ONC while CC and CCP overlap with each other (shown in S3). Our best model *CNN_P* ($fc6$) has an AUC of 1 for ONC classification, and very high AUC (0.91 and 0.86) for CC and CCP classification (Fig. 9 top).

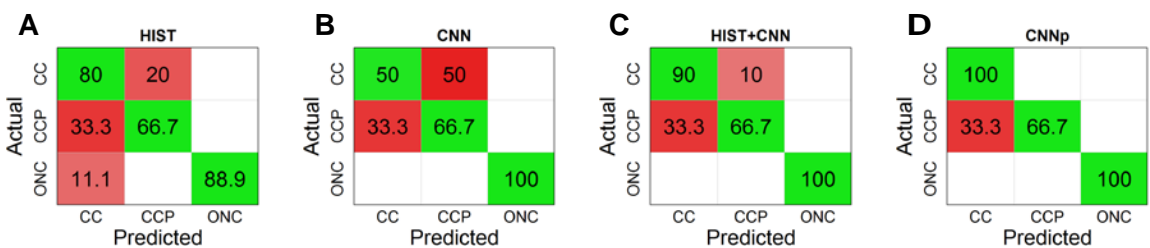

Figure 8: Selected confusion matrices of **A** *HIST*, **B** *CNN* ($fc8$), **C** *HIST* +*CNN* ($fc8$) and **D** *CNN_P* ($fc6$) classifiers. Percentage of predicted patients per class shown. While the whole image *CNN* is improved by histogram information, the patched *CNN_P* has access to more information and subtle details in the images.

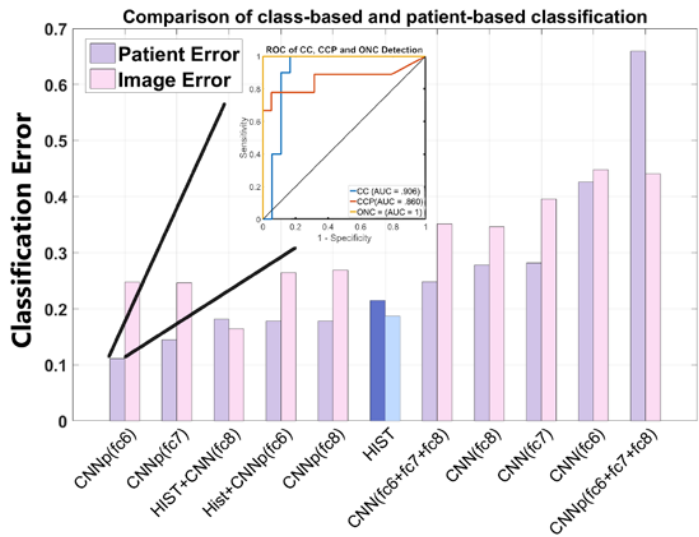

Figure 9: Result of LOPO-CV. Generally, histogram features *HIST* (middle) are better than *CNN* on whole images (right side), but worse than *CNN$_P$* on patches. The ROC of our best model *CNN$_P$* ($fc6$), has an AUC of 0.91 (CC), 0.86 (CCP) and 1 (ONC).

## Conclusion

The separation of RCC subtypes based on mitochondria quantification is an ongoing topic in cancer research, as mitochondria alterations seem to play a completely different role in RCC as in other types of cancer. We proposed a new solution to solve this difficult task based on structurally derived intensity histograms and state-of-the-art CNN feature vectors from three different layers of the well-studied AlexNet. We call these CNN features naïve, since the CNN has not been tuned to our dataset. Instead, the tuning process lies in the training of the random forest alone.

We compared these different approaches in a random forest cross-validation scenario and found CNNs on informative patches work best for this task. CNNs on whole images, however, may suffer from downscaling of the images (but still have moderate performace).

Finally, a classical histogram-based approach with hand-crafted features gives valuable results, even if worse than CNNs ($fc6$). This more complicated pipeline might offer three benefits: (i) the nuclei are only sampled on tissue (foreground), while the CNN would input the whole image, regardless of semantics (patching solves this to a certain extend, but introduces complexity), (ii) nucleus detection allows for explicit nucleus classification (Schueffler et al. (2010)), introducing further semantics, and (iii) Handcrafted features might be better interpretable than deep features.

We showed that a pre-trained CNN can already have highly predictive power on a highly specific histology problem even without CNN tuning. It might be expected that tuning further improves the classification.

# References

Yali Amit and Donald Geman. Shape quantization and recognition with randomized trees. *Neural Computation*, 9:1545–1588, 1997.

Pedram Argani, Cristina R. Antonescu, Peter B. Illei, Man Yee Lui, Charles F. Timmons, Robert Newbury, Victor E. Reuter, A. Julian Garvin, Antonio R. Perez-Atayde, Jonathan A. Fletcher, J. Bruce Beckwith, Julia A. Bridge, and Marc Ladanyi. Primary renal neoplasms with the aspl-tfe3 gene fusion of alveolar soft part sarcoma: A distinctive tumor entity previously included among renal cell carcinomas of children and adolescents. *The American Journal of Pathology*, 159(1):179 – 192, 2001.

Yaniv Bar, Idit Diamant, Lior Wolf, and Hayit Greenspan. Deep learning with non-medical training used for chest pathology identification. In *SPIE Medical Imaging*, 2015.

A. Rose Brannon, Anupama Reddy, Michael Seiler, Alexandra Arreola, Dominic T. Moore, Raj S. Pruthi, Eric M. Wallen, Matthew E. Nielsen, Huiqing Liu, Katherine L. Nathanson, Börje Ljungberg, Hongjuan Zhao, James D. Brooks, Shridar Ganesan, Gyan Bhanot, and W. Kimryn Rathmell. Molecular stratification of clear cell renal cell carcinoma by consensus clustering reveals distinct subtypes and survival patterns. *Genes & Cancer*, 1 (2):152–163, 2010. doi: 10.1177/1947601909359929.

Leo Breiman. Random forests. *Mach. Learn.*, 45:5–32, 2001. ISSN 0885-6125.

Qaiser Chaudry, Syed Hussain Raza, Andrew N. Young, and May D. Wang. Automated renal cell carcinoma subtype classification using morphological, textural and wavelets based features. *Journal of Signal Processing Systems*, 55(1):15–23, 2009. ISSN 1939-8115. doi: 10.1007/s11265-008-0214-6.

Dan C. Cireşan, Alessandro Giusti, Luca M. Gambardella, and Jürgen Schmidhuber. Mitosis Detection in Breast Cancer Histology Images with Deep Neural Networks. In *MICCAI*, pages 411–418. 2013. ISBN 978-3-642-40762-8 978-3-642-40763-5.

Angel Alfonso Cruz-Roa, John Edison Arevalo Ovalle, Anant Madabhushi, and Fabio Augusto González Osorio. A Deep Learning Architecture for Image Representation, Visual Interpretability and Automated Basal-Cell Carcinoma Cancer Detection. In *MICCAI*, pages 403–410. 2013. ISBN 978-3-642-40762-8 978-3-642-40763-5.

Caleb F. Davis, Christopher J. Ricketts, Min Wang, Lixing Yang, Andrew D. Cherniack, Hui Shen, Christian Buhay, Hyojin Kang, Sang Cheol Kim, Catherine C. Fahey, Kathryn E. Hacker, et al. The Somatic Genomic Landscape of Chromophobe Renal Cell Carcinoma. *Cancer Cell*, 26:319–330, 2014. ISSN 1535-6108.

Thomas J. Fuchs and Joachim M. Buhmann. Computational Pathology: Challenges and Promises for Tissue Analysis. *Computerized Medical Imaging and Graphics*, 35(7–8): 515–530, 2011. ISSN 0895-6111.

Kunihiko Fukushima. Neocognitron: A self-organizing neural network model for a mechanism of pattern recognition unaffected by shift in position. *Biological Cybernetics*, 36: 193–202, 1980. ISSN 0340-1200, 1432-0770.

A. Giusti, C. Caccia, D. C. Cireşari, J. Schmidhuber, and L. M. Gambardella. A comparison of algorithms and humans for mitosis detection. In *2014 IEEE 11th International Symposium on Biomedical Imaging (ISBI)*, pages 1360–1363, 2014.

A. Ari Hakimi, Satish K. Tickoo, Anders Jacobsen, Judy Sarungbam, John P. Sfakianos, Yusuke Sato, Teppei Morikawa, Haruki Kume, Masashi Fukayama, Yukio Homma, Ying-Bei Chen, Alexander I. Sankin, Roy Mano, Jonathan A. Coleman, Paul Russo, Seishi Ogawa, Chris Sander, James J. Hsieh, and Victor E. Reuter. TCEB1-mutated renal cell carcinoma: a distinct genomic and morphological subtype. *Mod Pathol*, 28(6):845–853, June 2015. ISSN 0893-3952.

Yangqing Jia, Evan Shelhamer, Jeff Donahue, Sergey Karayev, Jonathan Long, Ross Girshick, Sergio Guadarrama, and Trevor Darrell. Caffe: Convolutional architecture for fast feature embedding. *arXiv preprint arXiv:1408.5093*, 2014.

Shilpy Joshi, Denis Tolkunov, Hana Aviv, Abraham A. Hakimi, Ming Yao, James J. Hsieh, Shridar Ganesan, Chang S. Chan, and Eileen White. The Genomic Landscape of Renal Oncocytoma Identifies a Metabolic Barrier to Tumorigenesis. *Cell Reports*, 13, 2015. ISSN 2211-1247.

P. Kontschieder, M. Fiterau, A. Criminisi, and S. Rota Bulo'. Deep neural decision forests. In *Intl. Conf. on Computer Vision (ICCV), Santiago, Chile*, December 2015.

Alex Krizhevsky, Ilya Sutskever, and Geoffrey E. Hinton. Imagenet classification with deep convolutional neural networks. In *Advances in Neural Information Processing Systems 25*, pages 1106–1114. 2012.

S. Kullback and R. A. Leibler. On Information and Sufficiency. *The Annals of Mathematical Statistics*, 22:79–86, 1951. ISSN 0003-4851.

Quoc V. Le, Ju Han, Joe W. Gray, Paul T. Spellman, Alexander Borowsky, and Bahram Parvin. Learning invariant features of tumor signatures. In *Biomedical Imaging (ISBI), 2012 9th IEEE International Symposium on*, pages 302–305, 2012.

Mingsheng Long, Yue Cao, Jianmin Wang, and Michael I. Jordan. Learning transferable features with deep adaptation networks. *arXiv:1502.02791v2 [cs.LG]*, 2015.

David Meierhofer, Johannes A. Mayr, Ulrike Foetschl, Alexandra Berger, Klaus Fink, Nikolaus Schmeller, Gerhard W. Hacker, Cornelia Hauser-Kronberger, Barbara Kofler, and Wolfgang Sperl. Decrease of mitochondrial DNA content and energy metabolism in renal cell carcinoma. *Carcinogenesis*, 25, 2004. ISSN 0143-3334, 1460-2180.

H. Nilsson, D. Lindgren, A. Mandahl Forsberg, H. Mulder, H. Axelson, and M. E. Johansson. Primary clear cell renal carcinoma cells display minimal mitochondrial respiratory capacity resulting in pronounced sensitivity to glycolytic inhibition by 3-Bromopyruvate. *Cell Death & Disease*, 6, 2015.

S. H. Raza, Y. Sharma, Q. Chaudry, A. N. Young, and M. D. Wang. Automated classification of renal cell carcinoma subtypes using scale invariant feature transform. In *2009 Annual International Conference of the IEEE Engineering in Medicine and Biology Society*, pages 6687–6690, Sept 2009. doi: 10.1109/IEMBS.2009.5334009.

Ed Reznik, Martin L. Miller, Yasin Şenbabaoğlu, Nadeem Riaz, Judy Sarungbam, Satish K. Tickoo, Hikmat A. Al-Ahmadie, William Lee, Venkatraman E. Seshan, A. Ari Hakimi, and Chris Sander. Mitochondrial DNA copy number variation across human cancers. *eLife*, 5, 2016. ISSN 2050-084X.

Stephen M Rohan, Yonghong Xiao, Yupu Liang, Maria E Dudas, Hikmat A Al-Ahmadie, Samson W Fine, Anuradha Gopalan, Victor E Reuter, Marc K Rosenblum, Paul Russo, and Satish K Tickoo. Clear-cell papillary renal cell carcinoma: molecular and immunohis-tochemical analysis with emphasis on the von Hippel-Lindau gene and hypoxia-inducible factor pathway-related proteins. *Mod Pathol*, 24(9):1207–1220, 2011. ISSN 0893-3952.

A. C. Ruifrok and D. A. Johnston. Quantification of histochemical staining by color decon-volution. *Anal. Quant. Cytol. Histol.*, 23:291–299, 2001.

Judy Sarungbam, Ed Reznik, A Ari Hakimi, A Bialik, S. J. Sirintrapun, Hikmat A Al-Ahmadie, Anuradha Gopalan, Samson W Fine, Ying-Bei Chen, Chris Sander, Victor E Reuter, and Satish K Tickoo. Clear Cell Papillary Renal Cell Carcinoma Shows Marked Depletion of Mitochondrial Content: A Comparative Differential Diagnostic Study. *Mod Pathol*, (S2):260–261, February 2016.

Peter J. Schueffler, Thomas J. Fuchs, Cheng Soon Ong, Volker Roth, and Joachim M. Buhmann. Computational TMA Analysis and Cell Nucleus Classification of Renal Cell Carcinoma. In *DAGM*, 2010. ISBN 3-642-15985-0 978-3-642-15985-5.

Peter J. Schüffler, Thomas J. Fuchs, Cheng S. Ong, Peter Wild, and Joachim M. Buhmann. TMARKER: A Free Software Toolkit for Histopathological Cell Counting and Staining Estimation. *Journal of Pathology Informatics*, 4, 2013.

Amit Shah, Sailesh Conjeti, Nassir Navab, and Amin Katouzian. Deeply learnt hashing forests for content based image retrieval in prostate mr images. volume 9784, pages 978414–978414–6, 2016.

Tao Shi, David Seligson, Arie S. Belldegrun, Aarno Palotie, and Steve Horvath. Tumor classification by tissue microarray profiling: random forest clustering applied to renal cell carcinoma. *Mod Pathol*, 18(4):547–557, Oct 2004. ISSN 0893-3952.

The Cancer Genome Atlas Research Network. Comprehensive molecular characterization of clear cell renal cell carcinoma. *Nature*, 499(7456):43–49, 2013. ISSN 0028-0836.

S. K. Tickoo, M. W. Lee, J. N. Eble, M. Amin, T. Christopherson, R. J. Zarbo, and M. B. Amin. Ultrastructural observations on mitochondria and microvesicles in renal oncocytoma, chromophobe renal cell carcinoma, and eosinophilic variant of conventional (clear cell) renal cell carcinoma. *The American Journal of Surgical Pathology*, 24:1247–1256, 2000. ISSN 0147-5185.

Mitko Veta, Paul J. van Diest, Stefan M. Willems, Haibo Wang, Anant Madabhushi, Angel Cruz-Roa, Fabio Gonzalez, Anders B. L. Larsen, et al. Assessment of algorithms for mitosis detection in breast cancer histopathology images. *Medical Image Analysis*, 20: 237–248, 2015. ISSN 1361-8415.

Haibo Wang, Angel Cruz-Roa, Ajay Basavanhally, Hannah Gilmore, Natalie Shih, Mike Feldman, John Tomaszewski, Fabio Gonzalez, and Anant Madabhushi. Cascaded ensemble of convolutional neural networks and handcrafted features for mitosis detection. In *Proc. SPIE 9041, Medical Imaging 2014: Digital Pathology*, 2014.

Fang-Cheng. Yeh, Anil. Parwani, Liron. Pantanowitz, and Chien. Ho. Automated grading of renal cell carcinoma using whole slide imaging. *Journal of Pathology Informatics*, 5 (1):23, 2014. doi: 10.4103/2153-3539.137726.

Jason Yosinski, Jeff Clune, Anh Nguyen, Thomas Fuchs, and Hod Lipson. Understanding Neural Networks Through Deep Visualization. *arXiv preprint arXiv:1506.06579*, 2015.

# Deep vs. Flat Features for Mitochondria-based Renal Cell Carcinoma Subtyping - Supplement

## 1. Tissue Microarray

Figure 10 shows the tissue micro array where the individual spots have been identified and numbered automatically. 96 RCC patients with specimen triplicates are present, of which 28 could be used for this study. A large part of these patients have unclassified tumors. The diagnosis of the patients was done by trained pathologists on the underlying whole tissue slides. The TMA cores have been selected by pathologists as typical representative sites of the corresponding cancer.

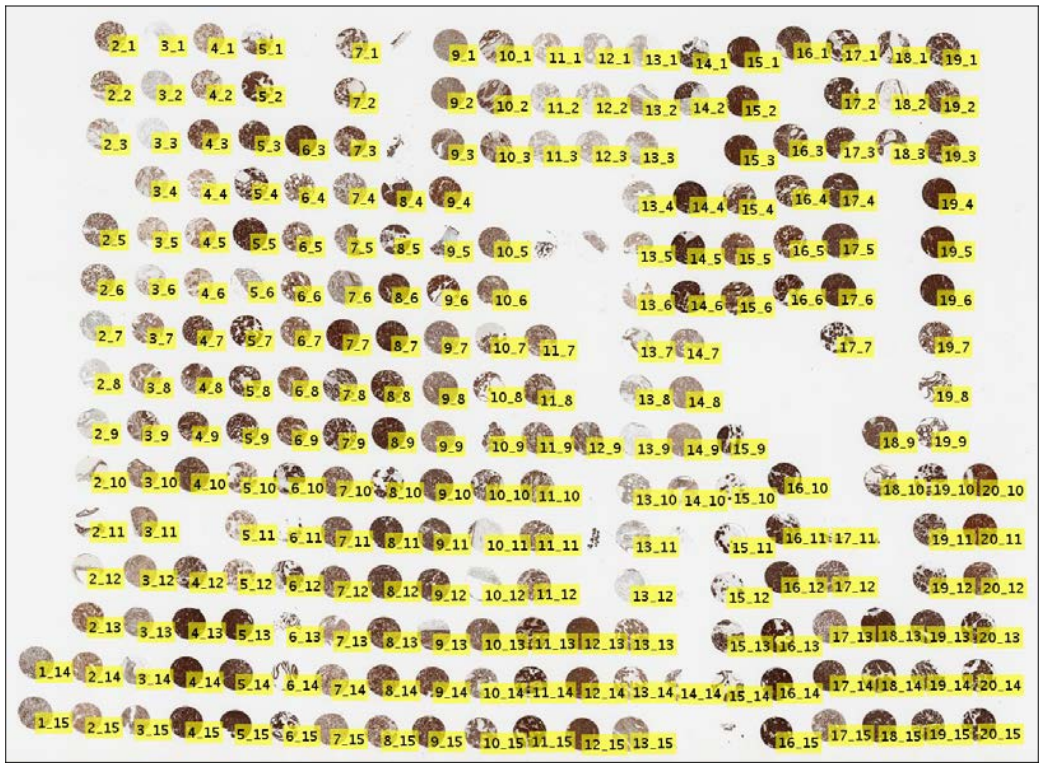

Figure 10: Overview of the tissue microarray of 92 RCC patients. 28 of these patients were diagnosed by the interesting RCC subtypes clear cell (CC), clear cell papillary (CCP) or oncocytoma (ONC). Each patient has specimen triplicates from primary tumor tissue, although individual tissue cores might be destroyed during preparation.

## 2. Number of Trees for Classification

We explored the random forest models' number of trees affecting the classification performance during cross-validation. The classification error does not decrease significantly after 50 trees (see Fig 11). Still, the computation time increases (especially for the CNN features, data not shown). We chose 50 trees to be consistently fixed for all for our experiments as a empirically good trade-off of performance and computation time. E.g. the training of one random forest on $CNN_P$ fc6 layer (4096 features, 1845 samples (patches)), takes 4.0 minutes of a standard desktop PC, with 50 trees. With 100 trees, the computation time takes 8.8 minutes.

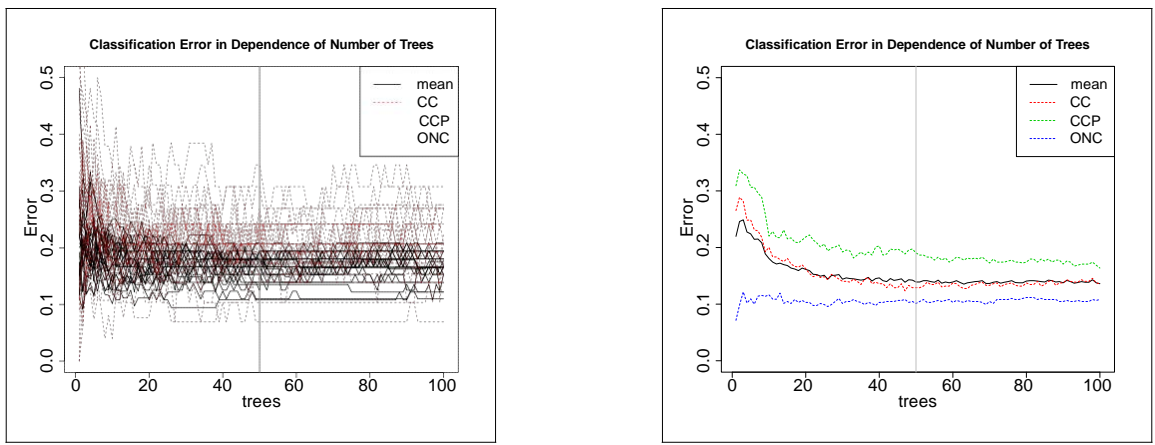

Figure 11: Classification error of a LOPO-CV in dependence of the number of trees in the forest. Classification accuracy per class shown. **Left:** LOPO-CV of *HIST* (517 features), **Right:** One fold of $CNN_P$ (4096 features). We chose 50 trees for all experiments (gray vertical line).

## 3. Histogram Clustering

To explore the potential of histograms to separate the cases, we calculated the Kullback-Leibler divergence $KL$ Kullback and Leibler (1951) between all histograms and the mean histograms. As we do not have a ground truth distribution but only observed distributions, we symmetrize $KL$ between two histograms $I$ and $J$ by

$$KL_{sym} = \frac{KL(I, J) + KL(J, I)}{2}$$

As we hypothesize the histograms to be similar in each class, we expect the divergence to be small within a subtype and larger between the classes. The multidimensional scaling of the dissimilarity matrix, which is shown in Figure 12, confirms that hypothesis as the Kullback-Leibler intra-class divergence is smaller than the inter-class divergence. Still, certain overlap of the classes indicates that individual cases from different classes show similar histograms. Also here, we observe the CC class to be clustered in between CCP and ONC.

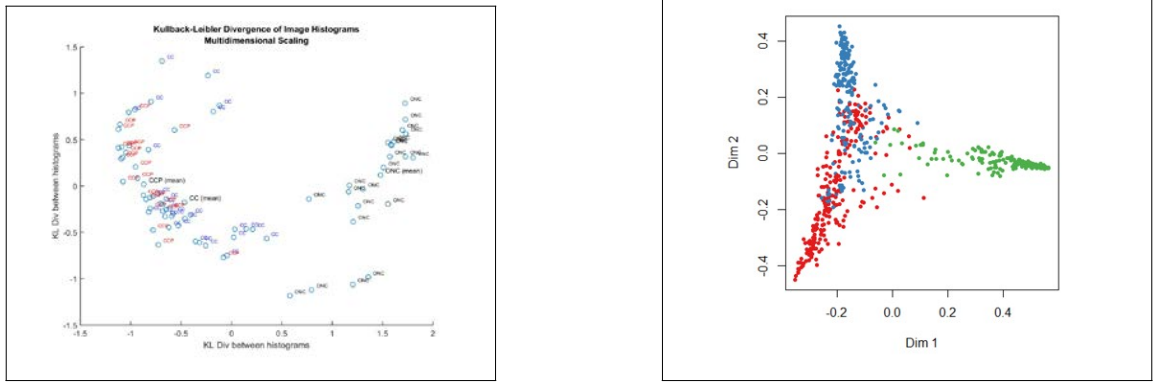

Figure 12: **Left:** Multidimensional scaling of the Kullback-Leibler divergence between all histograms of all 75 images plus the three mean histograms for the individual subtypes CC, CCP and ONC. Blue, CC; red, CCP; black, ONC. The intra-class divergence is generally smaller than the inter-class divergence, and classes can largely be separated. **Right:** Clustering of cRCC subtypes based on their mitochondrial staining by the multidimensional scaling of the CNN features. Red, CC. Blue, CCP. Green, ONC.